\documentclass{article}

\usepackage{natbib}
\setcitestyle{square,sort,comma,numbers}
\usepackage{arxiv}
\usepackage[utf8]{inputenc} 
\usepackage[T1]{fontenc}    
\usepackage{hyperref}       
\usepackage{url}            
\usepackage{booktabs}       
\usepackage{amsfonts}       
\usepackage{nicefrac}       
\usepackage{microtype}      
\usepackage{lipsum}

\usepackage{makeidx}
\usepackage{graphicx}
\usepackage[linesnumbered,ruled,vlined]{algorithm2e}

\newtheorem{PROPOSITION}{Proposition}
\newtheorem{proposition}[PROPOSITION]{Proposition}

\title{Dropout with Tabu Strategy for Regularizing Deep Neural Networks}

\author{Zongjie Ma$^1$\thanks{Corresponding author: zongjie.ma@griffithuni.edu.au ~~~~~~~~(Preprint and Work in progress.)}, Abdul Sattar$^1$, Jun Zhou$^1$, Qingliang Chen$^2$, Kaile Su$^{1}$\\
$^1$Institute for Integrated and Intelligent Systems, Griffith University, Brisbane, Australia\\
$^2$Department of Computer Science, Jinan University, Guangzhou, China\\
}

\begin{document}
\maketitle

\begin{abstract}
Dropout has proven to be an effective technique for regularization and preventing the co-adaptation of neurons in deep neural networks (DNN). It randomly drops units with a probability $p$ during the training stage of DNN. Dropout also provides a way of approximately combining exponentially many different neural network architectures efficiently. In this work, we add a diversification strategy into dropout, which aims at generating more different neural network architectures in a proper times of iterations. The dropped units in last forward propagation will be marked. Then the selected units for dropping in the current FP will be kept if they have been marked in the last forward propagation. We only mark the units from the last forward propagation. We call this new technique Tabu Dropout. Tabu Dropout has no extra parameters compared with the standard Dropout and also it is computationally cheap. The experiments conducted on MNIST, Fashion-MNIST datasets show that Tabu Dropout improves the performance of the standard dropout.
\end{abstract}


\section{Introduction}
With the vast amount of data and powerful hardware resources, deep learning algorithms have obtained high performance across many applications, such as computer vision \cite{DBLP:conf/nips/KrizhevskySH12}, natural language processing \cite{DBLP:conf/icml/CollobertW08}. DNN is one of the most popular showcases of deep learning algorithms. It contains multiple no-linear hidden layers, which can learn very complicated relationships in the data.
However, by building such a complex model, it is more prone to perfectly fit the training data while with less generalization for the real test data. This is called overfitting which is a major problem for Neural Networks. Many regularization methods have proposed to reduce overfitting including soft weight-sharing \cite{nowlan1992simplifying}, early stopping technique, $L_{1}$ and $L_{2}$ regularization.

The standard Dropout \cite{DBLP:journals/jmlr/SrivastavaHKSS14,DBLP:journals/corr/abs-1207-0580} is a widely used regularization technique that specific to DNN. During every iteration of training stage, this technique randomly shutdowns parts of neurons (hidden and visible) in a neural network with probability $p$ using samples from a Bernoulli distribution. So each iteration has a different ``thinned'' network. The resulting network is interpreted as a combination of these multiple ``thinned'' networks, which usually leads to a better generalization for unseen test data.

Dropout can be regarded as an ensemble method. As mentioned by \cite{DBLP:journals/jmlr/SrivastavaHKSS14}, the combination of multiple models is most helpful when the architectures of individual models are different from each other. Inspired by the Tabu Strategy successfully used in local search algorithms, in this work, we design a Tabu strategy to guide the Dropout to generate more different neural network architectures within a certain number of iterations. We call this new technique Tabu Dropout. From the second forward propagation of the training stage, a (0,1)-matrix is used to mark the dropped status of units in last forward propagation, which is called the Tabu list. More specifically, if
0 that unit is dropped, while if 1 the unit is kept. Then the units will not be allowed to drop if they have been marked as 0 in the Tabu list. The Tabu List is updated after each forward propagation and only stores the status of units from the last forward propagation. The reason for using this short-memory is to make a balance between the time-complexity and the diversification of neural networks. Tabu Dropout is computationally cheap and only has one parameter, the dropout rate $p$, as the standard dropout.

We carry out experiments to compare Tabu Dropout with the standard Dropout and AlphaDropout \cite{klambauer2017self} on the MNIST and Fashion-MNIST datasets. The experimental results show that Tabu Dropout outperforms the other two dropout techniques on these datasets. Especially, Tabu Dropout can achieve better performance in the early iterations, which will be useful for large and hard datasets.

\section{Preliminaries}
\subsection{Definitions and Notations}

Given a deep neural network $M$ with $L$ hidden layers,  indexed by $l \in \left \{1,2,3,...,L  \right \}$. Let $Z^{(l)}$ denote the input vectors of layer $l$. The output vectors from layer $l$ are denoted as $A^{(l)}$. As usual, $A^{(0)} = X$ is the input, $Z^{(L)}$ is the output the of neural network $M$. For layer $l$, $W^{(l)}$ is the weights matrix and $B^{(l)}$ is the bias vector.

\subsection{The Standard Dropout}
\begin{figure}[ht]
  \centering
  \begin{minipage}{.5\linewidth}
    \begin{algorithm}[H]
      \SetKwData{Left}{left}\SetKwData{This}{this}\SetKwData{Up}{up}
      \SetKwFunction{Union}{Union}\SetKwFunction{FindCompress}{FindCompress}
      \SetKwInOut{Parameters}{Parameters}\SetKwInOut{0dropout}{0dropout}
      \Parameters{a dropout rate $p \in (0,1)$, $f()$ is a activation function}

      \BlankLine

       \If{In Training Stage}{\label{train_stage}
                        $Mask^{(l)} \leftarrow Bernoulli(p)$;\label{mask}\\
                        $\hat{A}^{(l)} \leftarrow Mask^{(l)} *A^{(l)}$;\\
                        $\hat{A}^{(l)} \leftarrow \hat{A}^{(l)} / (1-p)$ ;\label{inverted-dropout}\\
                        $Z^{(l+1)} \leftarrow W^{(l+1)}*\hat{A}^{(l)}  + B^{(l+1)}$;\\
                        $A^{(l+1)} \leftarrow f(Z^{(l+1)})$;\label{}\\
                    }\ElseIf{In Testing Stage}{
                               $Z^{(l+1)} \leftarrow W^{(l+1)}*A^{(l)}  + B^{(l+1)}$;\\
                               $A^{(l+1)} \leftarrow f(Z^{(l+1)})$;\label{}\\
                               }

        \caption{Forward Propagation with Dropout}\label{standard_dropout}
    \end{algorithm}
  \end{minipage}
\end{figure}
Algorithm \ref{standard_dropout} describes how the dropout technique works in the forward propagation of a neural network. This algorithm can be divided into two parts: the training stage and the testing stage. In the training stage. The $Mask$, which is a (0,1)-matrix generated by samples from a Bernoulli distribution, is used to shutdown some  neurons by Line 3. After that, $\hat{A}^{(l)}$ is scaled by a factor of $1/(1-p)$, where $p$ is the probability of an unit to be zeroed. By doing this we are assuring that the result of the cost will still have the same expected value as without dropout, which is also named inverted dropout.

\subsection{Related Work}
\subsubsection{Variants of The Standard Dropout}
The standard Dropout \cite{DBLP:journals/corr/abs-1207-0580,DBLP:conf/nips/KrizhevskySH12} is successfully used into different kinds of neural networks models as a regularization technique to prevent overfitting. Some follow-up methods have been proposed to generally improve it. AlphaDropout \cite{klambauer2017self} aims at keeping the mean and variance to the original values of the input, which helps maintain the self-normalizing property. \cite{DBLP:conf/iccv/MorerioCVVM17} proposed a temporal dependent parameter to adjust dropout rate in the neural network. \cite{krueger2016zoneout} proposed Zoneout, which is a regularization technique for recurrent neural networks(RNNs)
It tries to reduce the variance across different dropout masks in the prediction stage of RNNs. Dropconnet \cite{DBLP:conf/icml/WanZZLF13}, is a  generalization of the standard Dropout, which randomly drops some weights. It has a better performance but slightly slower to train the neural networks compared with the standard dropout or Non-dropout. \cite{DBLP:conf/icml/RippelGA14} introduced Nested Dropout which is designed for unsupervised learning methods.

In this work we proposed a new Dropout technique called Tabu Dropout, which aims at guide the Dropout technique to diversity the sub neural network architectures within a certain number of iterations. Tabu Dropout also has a lower time-complexity and uses only one parameter (dropout rate $p$) as the same as the standard Dropout.

\section{Tabu Dropout}
\subsection{Motivation}
To illustrate the motivation of the standard dropout, \cite{DBLP:conf/nips/KrizhevskySH12} made a wonderful analogy with the sexual reproduction.

Further on, inspired by sexual reproduction may suffer from the inbreeding, we try to introduce some strategy to prevent the ``inbreeding'' when applying the standard dropout into training the neural networks. Tabu strategy \cite{DBLP:journals/informs/Glover89,DBLP:journals/informs/Glover90} is widely used in local search algorithm, which utilizes a memory structure (called tabu list) to forbid the local search to immediately return a previously visited candidate solution. We propose a new strategy which uses a (0,1)-matrix as the tabu list, which guides the standard Dropout to generate more diverse neural network structures.

\subsection{Model Description}
\subsection{The Standard Dropout}
\begin{figure}[ht]
  \centering
  \begin{minipage}{.5\linewidth}
    \begin{algorithm}[H]
      \SetKwData{Left}{left}\SetKwData{This}{this}\SetKwData{Up}{up}
      \SetKwFunction{Union}{Union}\SetKwFunction{FindCompress}{FindCompress}
      \SetKwInOut{Parameters}{Parameters}\SetKwInOut{0dropout}{0dropout}
      \Parameters{a dropout rate $p \in (0,1)$, $f()$ is a activation function}

      \BlankLine

       \If{In Training Stage}{\label{train_stage}
                        $Mask^{(l)} \leftarrow Bernoulli(p)$;\\
                        $Mask^{(l)} \leftarrow compare(Mask^{(l)}, Tabu)$; \label{times tabu}\\
                        $\hat{A}^{(l)} \leftarrow Mask^{(l)} * A^{(l)}$;\label{mask}\\
                        $\hat{A}^{(l)} \leftarrow \hat{A}^{(l)} / (1-p)$ ;\label{inverted-dropout}\\
                        $Z^{(l+1)} \leftarrow W^{(l+1)}*\hat{A}^{(l)}  + B^{(l+1)}$;\\
                        $A^{(l+1)} \leftarrow f(Z^{(l+1)})$;\label{}\\
                        $Tabu \leftarrow Mask^{(l)}$;\label{update tabu}\\
                    }\ElseIf{In Testing Stage}{
                               $Z^{(l+1)} \leftarrow W^{(l+1)}*A^{(l)}  + B^{(l+1)}$;\\
                               $A^{(l+1)} \leftarrow f(Z^{(l+1)})$;\label{}\\
                               }

        \caption{Forward Propagation with Tabu Dropout}\label{tabu_dropout}
    \end{algorithm}
  \end{minipage}
\end{figure}
Tabu Dropout technique is described in Algorithm \ref{tabu_dropout}. It is also only used in the training stage of the forward propagation. We use a (0,1)-matrix as the Tabu list to mark the drop status of each neural is the last forward propagation. The status of this unit is 0 if it has been dropped in lasted forward propagation and 1 otherwise. To be specific, when training a neural network, Tabu Dropout follows the standard Dropout in the first forward propagation and generates a $Tabu$ matrix assigned by the dropout mask.  The Dropout mask is (0,1)-matrix generated from a Bernoulli distribution. Then in the following forward propagation, we will make a comparison between $Mask$ and $Tabu$, if a unit has the same status with 0 in both $Mask$ and $Tabu$, the status of this unit will be changed into 1 after Line~\ref{times tabu}. That is, the units dropped in the last forward propagation will not be allowed to drop in the coming forward propagation. As a result, Tabu Dropout will guide the the standard dropout to diversity the sub neural network architectures within a certain number of iterations. In order to make a balance between diversity and time-complexity, the $Tabu$ list only marks the status form the last forward propagation, and the $Tabu$ list will be overrode by the new $Mask$ after each forward propagation in the training stage(Line ~\ref{update tabu}). By this way, Tabu Dropout only has a short memory for guiding the standard Dropout.

There are two advantages in Tabu Dropout:
\begin{itemize}
\item There is no extra parameter compared with the standard dropout.
\item Experimental results show that Tabu Dropout is very computationally cheap.
\end{itemize}

Give a fixed Dropout rate $p$ in Tabu Dropout. After using Tabu in each forward propagation, the actual dropout rate $\hat{p}$ with respect to all the units will be automatically adjusted according to the following Proposition~\ref{ajust-dropout-rate}.
\begin{proposition}\label{ajust-dropout-rate}
Let $n$ denotes the $n$th forward propagation, $p$ denotes the fixed dropout rate,  $\hat{p}_{n}$ denotes the dropout rate with respect to all the units in the $n$th forward propagation, we have:

~~~~~~~~~~~~~~~~~~$\hat{p}_{(n+1)} = (1-\hat{p}_{n})\times p$ \\
where $\hat{p}_{1}=p, n \in N$ and $n >= 1$
\end{proposition}

\section{Experiment}
In this section, we evaluate the Tabu Dropout in Multilayer Perceptrons (MLP),which is a full-connected neural network, for classification problems on two different datasets. Tabu Dropout outperforms the standard Dropout and AlphaDropout on these two datasets.

\subsection{Datasets}
We conduct the experiments on the two datasets, the details of these datasets are as follows.

\begin{figure}
  \includegraphics[width=\linewidth]{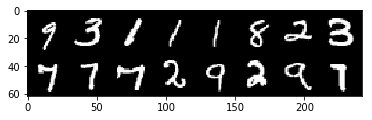}
  \caption{6 samples from MNIST dataset}
  \label{fig:mnist-sample}
\end{figure}

\textbf{MINST}\footnote{http://yann.lecun.com/exdb/mnist/} \cite{lecun1998gradient}: It is a classic handwritten digits dataset consists of 28$\times$28 grayscale images of the 10 digits(from 0 to 9). There are 60,000 training images and along with 10,000 testing images. MINST is widely used in machine learning and computer vision. Figure~\ref{fig:mnist-sample} shows some samples from MNIST.

\begin{figure}
  \includegraphics[width=\linewidth]{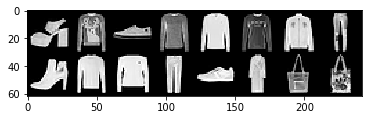}
  \caption{samples from Fashion-MNIST dataset}
  \label{fig:fasion-mnist-sample}
\end{figure}

\textbf{Fashion-MNIST}\footnote{https://github.com/zalandoresearch/fashion-mnist}\cite{xiao2017/online}: This is a dataset derived from the assortment on Zalando’s  website \footnote{http://www.zalando.com}. There are 60,000 examples in the training set and 10,000 examples in the testing set. Each example is a grayscale image consisting of 28 $\times$ 28 pixels and with a label from 10 classes (T-shirt/top, Trouser, Pullover, Dress, Coat, Sandal, Shirt, Sneaker, Bag, Ankle boot). There are some samples from Fashion-MNIST dataset shown in Figure \ref{fig:fasion-mnist-sample}.

\subsection{Experiment Setup}
We have implemented Tabu Dropout in Pytorch \cite{paszke2017automatic}, which is is a deep learning framework for fast, flexible experimentation. We use the standard Dropout layer and AlphaDropout layer directly in \texttt{torch.nn} from Pytorch. All the experiments are conducted on a 3.70GHz Intel core i7-8700K CPU, 32 GB RAM  and a TITAN Xp GPU under Ubuntu 18.04.1 LTS.
\subsection{Results on MNIST}
We use a MLP model, which is fully-connected, to evaluate the performances of Tabu Dropout, the standard Dropout, AlphaDropout and Non-Dropout. The model structure is: \texttt{LINEAR(1024 units)->RELU->LINEAR(1024 units)->RELU->LINEAR(10 units)->SIGMOID}. Different dropout strategies are added between the two hidden layers (each has 1024 units).

\begin{figure}
  \includegraphics[width=\linewidth]{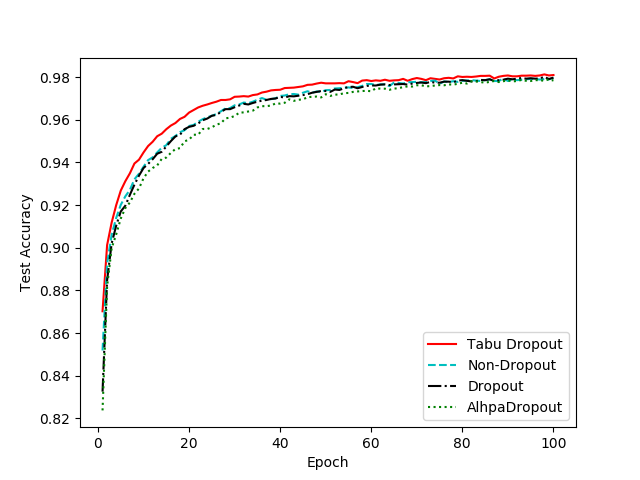}
  \caption{Accuracy rate of Model MLP with  Tabu Dropout, Dropout, AlphaDropout, Non-Dropout on test data of MNIST}
  \label{fig:mnist-results}
\end{figure}

Figure \ref{fig:mnist-results} shows the results of the accuracy rate of the MLP model with Tabu Dropout, the standard Dropout, Non-Dropout and AlphaDropout on MNIST test data. The dropout rate $p$ is fixed to 0.5 for both Tabu Dropout and the standard Dropout. For AlphaDropout, the dropout rate is set to 0.05 recommended by \cite{klambauer2017self}, and the experimental results show that AlphaDropout leads to a worse performance with $p$ = 0.5. The model with different dropout strategies or Non-dropout has been trained for 100 epochs, with learning rate 0.01 and batch size 512. The model with Tabu Dropout outperforms the other two models across all the trained epochs. The Non-dropout model goes to overfitting after a certain epochs.

\subsection{Results on Fashion-MNIST}
The same MLP model is used to evaluate the performance of Tabu Dropout, Dropout,  Non-Dropout and AlphaDropout on Fashion-MNIST dataset.

\begin{figure}
  \includegraphics[width=\linewidth]{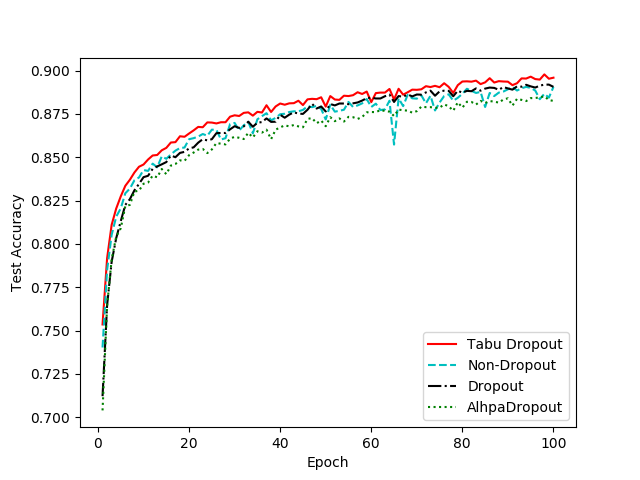}
  \caption{Accuracy rate of Model MLP with Tabu Dropout, Dropout, AlphaDropout, Non-Dropout on test data of Fashion-MNIST}
  \label{fig:Fashion-mnist-results}
\end{figure}

The experimental results are shown is Figure \ref{fig:Fashion-mnist-results}. We observe that Tabu Dropout always achieves higher accuracy on the 10,000 test dataset. The model with the standard Dropout falls behind at first and then gradually outperforms the model without Dropout. The reason is that, the model without dropout suffers from overfitting after a certain iterations for training. On the other hand, Tabu Dropout outperforms the Non-Dropout even at a early stage of training, which reveals that Tabu Dropout generates better sub neural network architectures for this dataset.

\section{Conclusions}
In this work we proposed Tabu Dropout for preventing overfitting in training the neural networks. Compared with the standard Dropout, it has a short memory of dropped units happened in the last dropout out. The units with a Tabu status of 0 will not allowed to drop in the coming forward propagation. The experimental results on two standard datasets show that Tabu Dropout outperforms the standard dropout, Alphadropout and Non-dropout in a same MLP model. Tabuout has a low time-complexity and keep the same parameter with the standard Dropout.

\section{Future Work}
We will use different deep neural network models such as convolutional neural networks (CNN), recurrent neural networks(RNN) and larger datesets such Imagenet\cite{imagenet_cvpr09} to evaluate the performance of Tabu Dropout in future.

\bibliographystyle{unsrt}
\bibliography{references}

\end{document}